\documentclass[11pt]{article}

\usepackage{acl}

\usepackage{times}
\usepackage{latexsym}
\usepackage{makecell}

\usepackage[T1]{fontenc}

\usepackage[utf8]{inputenc}

\usepackage{microtype}

\usepackage{inconsolata}

\usepackage{graphicx}

\usepackage{hyperref}

\title{EXL Health AI Lab at MEDIQA-WV 2025: Mined Prompting and Metadata-Guided Generation for Wound Care Visual Question Answering}

\author{
\textbf{Bavana Durgapraveen, Sornaraj Sivasankaran,} \\ 
\textbf{Abhinand Balachandran, Sriram Rajkumar} \\
EXL Service\\
\texttt{\{bavana.durgapraveen, sriram.rajkumar, abhinand.b, e.sivasankaran\}@exlservice.com}
}

\begin{document}
\nolinenumbers
\pagestyle{empty} 
\maketitle
\begin{abstract}
The rapid expansion of asynchronous remote care has intensified provider workload, creating demand for AI systems that can assist clinicians in managing patient queries more efficiently. The MEDIQA-WV 2025 shared task addresses this challenge by focusing on generating free-text responses to wound care queries paired with images. In this work, we present two complementary approaches developed for the English track. The first leverages a mined prompting strategy, where training data is embedded and the top-k most similar examples are retrieved to serve as few-shot demonstrations during generation. The second approach builds on a metadata ablation study, which identified four metadata attributes that consistently enhance response quality. We train classifiers to predict these attributes for test cases and incorporate them into the generation pipeline, dynamically adjusting outputs based on prediction confidence. Experimental results demonstrate that mined prompting improves response relevance, while metadata-guided generation further refines clinical precision. Together, these methods highlight promising directions for developing AI-driven tools that can provide reliable and efficient wound care support.
\end{abstract}

\section{Introduction}

The proliferation of remote patient care, accelerated by telehealth technologies, has transformed how patients and providers interact. Patients can now communicate asynchronously through secure portals, often submitting free-text messages and images for clinical review. While this model greatly improves accessibility and continuity of care, it has also generated new challenges for healthcare systems. Providers face an ever-growing volume of digital queries, creating what has been termed the “inbox burden” (\cite{sinsky-etal-2024-inbox}). This constant stream of patient messages can delay response times, reduce clinical efficiency, and contribute to physician burnout.
Artificial intelligence (AI)–based natural language generation offers a promising strategy to alleviate this workload. By producing high-quality draft responses to patient messages, such systems can streamline communication workflows, reduce repetitive documentation tasks, and allow clinicians to devote more time to complex decision-making. Previous work has shown that retrieval-augmented generation (RAG) methods (\cite{lewis-etal-2020-rag}; \cite{gao-etal-2023-scaling}) and clinical domain adaptation of large language models (LLMs) (\cite{singhal-etal-2023-llms}; \cite{lehman-etal-2023-gpt4}) can substantially improve the quality and reliability of AI-generated text in medical settings. However, applying these models in specialized areas such as wound care remains relatively unexplored.
Wound care presents unique challenges for automated response generation. Accurate assessment often depends on both visual attributes (e.g., wound type, tissue appearance, exudate characteristics) and textual context (e.g., patient-reported symptoms, history of treatment). This multimodal nature requires systems that can integrate visual and textual signals to produce clinically appropriate outputs. The MEDIQA-WV 2025 shared task \cite{yim-etal-2025-mediqawv} directly addresses this gap by providing a benchmark for generating free-text responses to patient wound care queries that include both text and images. The task advances prior MEDIQA challenges (\cite{benabacha-etal-2021-medicqa}; \cite{yim-etal-2023-mediqa}) by focusing on asynchronous, visually grounded care scenarios, thereby moving closer to real-world clinical applications.
In this paper, we present the work, developed for the English track of MEDIQA-WV 2025. Our central hypothesis is that generic, end-to-end vision-language models may lack the domain-specific grounding required for wound care queries. 
To address this, we investigate two complementary approaches: 
\begin{enumerate}
    \item  \textbf{A mined few-shot prompting} strategy, where the system retrieves clinically similar examples from the training data to guide generation, and 
    \item  \textbf{A metadata-guided generation} strategy, where structured wound attributes predicted by classifiers are incorporated into the generation process.
\end{enumerate}


\section{Shared Task and Dataset}

The MEDIQA-WV 2025 shared task focuses on wound care visual question answering (VQA), where the goal is to generate clinically coherent responses to patient queries about wounds by leveraging both wound images and textual inputs. The task is built on the recently introduced WoundcareVQA dataset (\cite{yim-etal-2025-mediqawv}), which consists of approximately 500 multilingual patient queries (English and Chinese)(Table \ref{tab:dataset_stats}). Each query is paired with one or two wound images and multiple expert-authored responses, enabling a multimodal setup that requires both visual and linguistic reasoning.In addition to raw queries and expert responses, each case is annotated with structured metadata covering clinically relevant wound attributes.These attributes serve as a rich source of metadata, covering aspects such as:

\begin{itemize}
    \item \textbf{Anatomic Location} (e.g., lower leg, abdomen, fingernail)
    \item \textbf{Wound Type} (e.g., surgical, traumatic, pressure ulcer)
    \item \textbf{Wound Thickness} (e.g., superficial, full thickness)
    \item \textbf{Tissue Color} (e.g., pink, red and moist, black)
    \item \textbf{Drainage Type} (e.g., serous, serosanguinous)
    \item \textbf {Drainage Amount} (e.g., scant, minimal, moderate)
    \item \textbf{Signs of Infection}
\end{itemize}

\begin{table*}[htbp!]
  \centering
  \begin{tabular}{lcccccc}
    \hline
    \textbf{Split} & 
    \makecell{\textbf{Cases}} & 
    \makecell{\textbf{Images}} & 
    \makecell{\textbf{Responses}} & 
    \makecell{\textbf{Responses} \\ \textbf{per Case}} & 
    \makecell{\textbf{Avg. Query} \\ \textbf{Length}} & 
    \makecell{\textbf{Avg. Response} \\ \textbf{Length}} \\
    \hline
    Training   & 279 & 449 & 279 & 1 & 44 words & 29 words \\
    Validation & 105 & 147 & 210 & 2 & 47 words & 36 words \\
    Test       & 93  & 152 & 279 & 3 & 52 words & 47 words \\
    
    \hline
    Total      & 477 & 748 & 768 & -- & -- & -- \\
    \hline
  \end{tabular}
  \caption{Dataset statistics across Training, Validation, and Test splits.}
  \label{tab:dataset_stats}
\end{table*}

An important characteristic of this dataset is the variability in inter-annotator agreement (IAA) across wound attributes. For example, wound type (1.0), tissue color (0.97), and infection (0.97) achieved near-perfect agreement, suggesting these features are well-defined and consistently identified by clinicians. In contrast, anatomic location (0.81), drainage amount (0.86), and wound thickness (0.89) show relatively lower agreement, highlighting attributes that are either more subjective or context-dependent. These differences emphasize that while certain wound features provide highly reliable signals for model training and evaluation, others introduce ambiguity that must be accounted for in system design and assessment.
This combination of free-text responses and structured wound attributes makes the dataset uniquely suited for hybrid approaches that combine classification and generation, and provides an opportunity to evaluate how multimodal systems handle both objective and subjective aspects of wound care reasoning.

\section{Related Works}
\begin{figure*}[t] \includegraphics[width=1\linewidth]{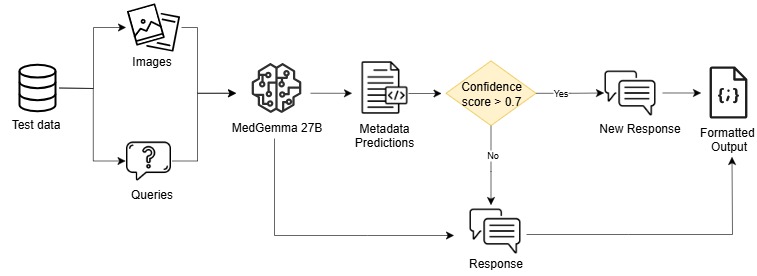} \hfill 
\caption {The diagram illustrates a workflow where test data (images and queries) are processed by a model to predict metadata, validated by confidence scoring, and transformed into formatted output responses.}
\end{figure*}

\begin{figure*}[t] \includegraphics[width=1\linewidth]{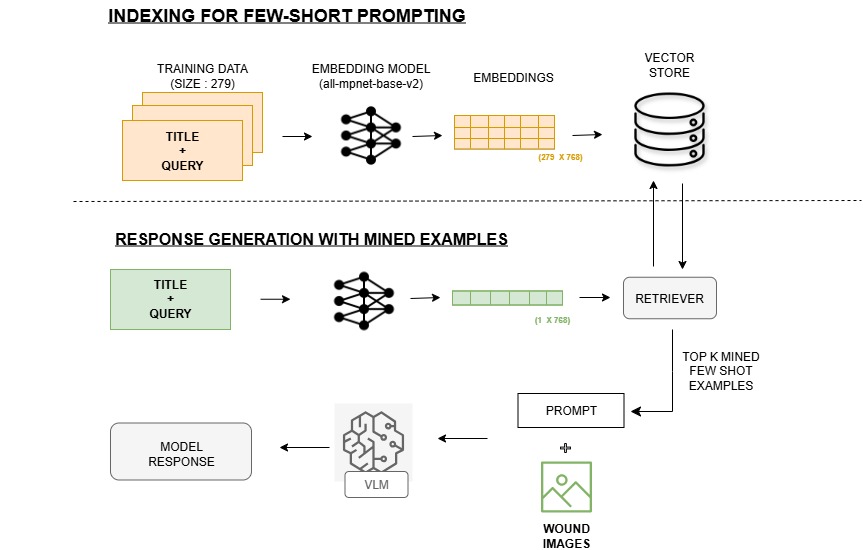} \hfill 
\caption {The Diagram illustrates overview of the mined few-shot prompting strategy with semantic similarity-based example retrieval and model-specific context optimization.}
\end{figure*}
In recent years, multimodal machine learning has gained considerable traction in healthcare applications, particularly with the rise of large multimodal models. Several open-source initiatives have pushed this field forward, including LLaVA-Med \cite{li-etal-2023-llava-med} and ELIXR \cite{xu-etal-2023-elixr}. The latter is especially notable for exploring CLIP-inspired training strategies, which closely align with the objectives of our work. Much of the current research has centered on radiology and other imaging-heavy specialties, while dermatology has received relatively limited attention. Notable early studies, such as Cirone et al. (2024), demonstrated that GPT-4o can distinguish melanoma from benign skin lesions with high reliability. However, this type of binary diagnostic task is substantially more constrained than the open-ended dermatology question answering examined in the present shared task, where queries and conditions may extend beyond the model’s training distribution. The difficulty of this broader problem is evident in our findings: although our system achieved only moderate overall accuracy, it nonetheless ranked first in the competition. This outcome underscores both the progress achieved and the significant challenges that remain in developing robust multimodal systems for dermatology. These results point to several important directions for future research, including evaluation frameworks that better reflect clinical utility \cite{kelly-etal-2019-ai-impact} and methods to enhance multimodal reasoning beyond narrow diagnostic endpoints.
Our methodology builds on two key ideas. First, structured attribute prediction is a well-established strategy in computer vision for grounding model decisions in interpretable features (\cite{russakovsky-etal-2015-imagenet}; \cite{zhang-etal-2023-metadata}). Second, our prompt mining strategy draws from retrieval-augmented generation (RAG) and in-context learning, where supplementing prompts with carefully selected examples has been shown to markedly improve large language model performance on domain-specific tasks (\cite{lewis-etal-2020-rag}; \cite{gao-etal-2023-scaling}; \cite{khandelwal-etal-2020-nnmt}).

\section{Methodology}
Our system is a pipeline designed to maximize the use of both structured and unstructured information available in the dataset. It leverages two powerful models and is orchestrated through two distinct approaches.

\subsection{Model Description}
MedGemma (27B):MedGemma (27B Multimodal), from Google’s Gemma 3 family, is a large language model specialized for medical contexts. Crucially, its multimodal variants integrate MedSigLIP—a 400-million parameter dual-tower vision–language encoder (SigLIP-based), pre-trained on diverse medical imaging data (e.g., dermatology, radiology, pathology. MedSigLIP powers the visual understanding in MedGemma, allowing the model to reason across modalities. While we treat MedGemma separately from InternVL, its built-in image encoder makes it a strong alternative for end-to-end medical image understanding and free-text clinical response generation, particularly when vision and language reasoning need seamless integration.

InternVL 3 (38B): InternVL-3 38B is an advanced multimodal large language model (MLLM) that demonstrates superior multimodal perception and reasoning capabilities compared to its predecessor InternVL 2.5The architecture follows the "ViT-MLP-LLM" paradigm with pixel unshuffle operations that reduce visual tokens to one-quarter of the original. The model extends multimodal capabilities to encompass tool usage, GUI agents, industrial image analysis, and 3D vision perception. InternVL3-38B achieves competitive performance with leading models like GPT-4o on multimodal benchmarks, making it particularly suitable for complex visual-linguistic tasks such as medical image analysis and wound care assessment applications requiring precise visual-textual integration.

\subsection{Approach 1: Metadata ablation study and conditional metadata prediction for response generation}

The goal of this study was to identify which clinical features had the most significant impact on response quality and to develop a strategy for leveraging them effectively.
First, we systematically evaluated the importance of each of the seven metadata categories provided in the dataset. By removing one category at a time from a full-context prompt and measuring the resulting drop in the deltaBLEU score, we quantified the contribution of each feature. This empirical analysis, combined with an examination of the dataset's inter-annotator agreement (IAA) scores, this score is a heuristic derived directly from the relaxed inter-annotator agreement scores provided with the dataset, which serves as a proxy for the reliability of a given category. Figure 1 illustartes the overview flow of metadata study approach
Based on these findings, we designed a two-stage pipeline centered on high-impact features:
\begin{enumerate}
    \item 	\textbf{Metadata Prediction}: For each instance in the test set, we use MedGemma (27B) to predict values for the four selected metadata categories. The task is framed as a few-shot classification problem where the model is prompted to select the most appropriate label from the predefined list based on the patient's query and a description of the images.
    \item \textbf{Confidence Score Assignment and Conditional Integration}: To account for the inherent ambiguity in clinical assessment, we assign a "confidence score" to each predicted metadata field. In the second stage of our pipeline, this predicted metadata and its confidence score are passed as context alongside the original query and images. We use a confidence threshold of 0.7 to determine how this information shapes the final response. If a metadata field's confidence is greater than or equal to 0.7, its predicted value is integrated into the prompt as a factual observation. If the confidence is lower, the prompt instructs the model to be cautious about that aspect, preventing overconfident and potentially incorrect advice.
\end{enumerate}

\subsection{Approach 2: Prompt Mining}
In this approach, we leverage a dynamic few-shot prompting strategy to generate clinically grounded annotated familiar responses by providing the model with familiar annotations from the training data. First, the training dataset is embedded using the all-mpnet-base-v2 sentence transformer, enabling efficient semantic similarity search. For each test query, the most relevant examples are retrieved from the training set and incorporated as few-shot examples in the prompt, allowing the model to learn from familiar patterns and annotations it encountered during training. Figure 2 illustrates the overview of prompt mining startegy approach. Through experimentation, we explored retrieval configurations ranging from top 5 to top 25 similar samples to determine the optimal context size. Specifically, we employed the InternVL3-38B model with the top 25 retrieved samples and the MedGemma-27B model with the top 5 retrieved samples, allowing each model to benefit from context sizes suited to its architecture. The enriched prompts, containing both the patient's query and carefully selected training examples, guide the models to produce accurate, coherent, and clinically appropriate responses.

\section{Evaluation Metrics}
System performance was evaluated using the official metrics of the shared task, which are designed for multi-reference free-text generation:

\textbf{deltaBLEU}: A variant of SacreBLEU developed for response generation, a case in which many diverse gold standard responses are possible (Galley et al., 2015). The metric incorporates humanannotated quality rating and assigns higher weights to n-grams from responses rated to be of higher quality. The authors have shown this method produces higher correlation with human rankings compared to previous BLEU metrics. In our system, we assign response weights according to four criteria: (a) if user expertise level is 4 or above (out of 9), (b) if user is formally validated as a medical doctor by the platform, (c) if the response answer is the most frequent answer, and (d) if the response answers the query completely. The former two were manually assigned to the validation and test sets by two NLP scientists. The test set was double-reviewed. Out of a 0.0-1.0 scale, if (d) is not met, the score is discounted to 0.9; for the other 3 criteria, 0.1 is discounted for every missing element to reach the final weight.

\textbf{BERTScore}: An embedding-based metric that measures the semantic similarity between the generated and reference texts.

\textbf{ROUGE-L}: A recall-oriented metric that measures the longest common subsequence.

\section{Results and Discussion}
\subsection{Performance Comparison}
Table \ref{tab:model_comparison_transposed} presents a comparative evaluation of different approaches across multiple metrics, including deltaBLEU and ROUGE-L for automated quality assessment, as well as DeepSeekV3, Gemini, and GPT-4o scores to capture model-specific performance. Additionally, Average Human Evaluation is reported to provide a subjective measure of overall quality. The results highlight that the MedGemma-27B (5-shot) approach achieves the highest deltaBLEU score (13.04), indicating strong alignment with reference outputs, while InternVL3-38B (25-shot) demonstrates competitive performance across human and model-based evaluations. Meanwhile, the Metadata Ablation Study serves as a baseline, showing moderate yet consistent results across all metrics. This comparison underscores the complementary role of automated and human evaluations in benchmarking advanced models. Top 2 results from the MEDIQA-WV 2025 shared task are from the two approaches we mentioned above InternVL-38B Mined few shot with 25 samples being in the first place \cite{yim-etal-2025-mediqawv}.

\begin{table*}[htbp]
    \centering
    \setlength{\tabcolsep}{6pt} 
    \renewcommand{\arraystretch}{1.2} 
    \begin{tabular}{cccc}
        \hline
        \textbf{Metric} & \textbf{Intern VL 38B} & \textbf{MedGemma 27B} & \makecell{\textbf{Metadata Study} \\ \textbf{MedGemma 27B}} \\
        \hline
        deltaBLEU            & 9.9152  & \textbf{13.0379} & 5.7015 \\
        ROUGE-1              & 0.7909  & 0.7118  & \textbf{0.8100} \\
        ROUGE-2              & \textbf{0.5613}  & 0.5128  & 0.5361 \\
        ROUGE-L              & \textbf{0.4561}  & 0.4517  & 0.4555 \\
        ROUGE-Lsum           & 0.4560  & \textbf{0.4572}  & 0.4553 \\
        BERTScore Mean-of-Mean & 0.6218  & 0.6188  & \textbf{0.6228} \\
        BERTScore Mean-of-Max  & 0.6690  & \textbf{0.6743}  & 0.6570 \\
        DeepSeekV3           & \textbf{0.6823}  & 0.6349  & 0.6070 \\
        Gemini               & \textbf{0.6452}  & 0.5914  & 0.6290 \\
        GPT-4o               & \textbf{0.7151}  & 0.6290  & 0.6667 \\
        \hline
        \textbf{Average}    & \textbf{0.4730}  & \textbf{0.4575}  & \textbf{0.4505} \\
        \hline
    \end{tabular}
    \caption{Performance comparison of Intern VL 38B, MedGemma 27B, and Metadata Study MedGemma 27B across multiple evaluation metrics.}
    \label{tab:model_comparison_transposed}
\end{table*}

\subsection{Metadata Ablation Study}
To identify the most clinically relevant features for wound assessment, we conducted a systematic ablation study examining the contribution of each metadata category. Table \ref{tab:metadata_ablation} demonstrates the impact of removing individual metadata components on model performance.

\begin{table*}[h!]
  \centering
  \begin{tabular}{lcc}
    \hline
    \textbf{System Configuration} & \textbf{deltaBLEU} & \textbf{Performance Drop} \\
    \hline
    All metadata classes         & 4.476 & -     \\
    Without metadata             & 3.786 & -0.690 \\
    Without infection            & 4.384 & -0.092 \\
    Without drainage type        & 4.254 & \textbf{-0.222} \\
    Without drainage amount      & 4.962 & +0.486 \\
    Without tissue color         & 4.021 & \textbf{-0.455} \\
    Without wound thickness      & 4.976 & +0.500 \\
    Without wound type           & 4.014 & \textbf{-0.462} \\
    Without anatomical location  & 3.960 & \textbf{-0.516} \\
    \hline
  \end{tabular}
  \caption{Metadata Ablation Study results showing deltaBLEU scores and performance drops when individual metadata components are removed.}
  \label{tab:metadata_ablation}
\end{table*}

The ablation results reveal that anatomical location (-0.516), wound type (-0.462), and tissue color (-0.455) cause the most significant performance degradation when removed, indicating their critical importance for accurate wound assessment. Conversely, removing wound thickness (+0.500) and drainage amount (+0.486) actually improved performance, suggesting these features may introduce noise or ambiguity in the current dataset context

\subsection{Inter-Annotator Agreement and Feature Selection}
The dataset exhibits considerable variability in inter-annotator agreement (IAA) across wound attributes, which directly correlates with their clinical utility. Wound type (1.0), tissue color (0.97), and infection (0.97) achieved near-perfect agreement, indicating these features are well-defined and consistently identified by clinicians. In contrast, anatomical location (0.81), drainage amount (0.86), and wound thickness (0.89) demonstrated lower agreement, highlighting attributes that are more subjective or context-dependent.
Based on the combined analysis of ablation study results and IAA scores, we selected anatomical location, wound type, drainage type, and tissue color as the most important metadata features for test data prediction. This selection strategy prioritizes features that either demonstrate high clinical impact (anatomical location, wound type, tissue color) or maintain reasonable reliability despite moderate IAA scores (drainage type: 0.92 IAA).
\begin{table*}[htbp]
    \centering
    \setlength{\tabcolsep}{8pt} 
    \renewcommand{\arraystretch}{1.2} 
    \begin{tabular}{cccc}
        \hline
        \textbf{Model} & \textbf{LLM as Judge} & \textbf{BERTScore Avg} & \textbf{ROUGE Avg} \\
        \hline
        InternVL38B          & 0.6808 & 0.6454 & 0.5661 \\
        MedGemma27B          & 0.6185 & 0.6465 & 0.5334 \\
        Metadata MedGemma27B & 0.6342 & 0.6399 & 0.5642 \\
        \hline
    \end{tabular}
    \caption{Evaluation of models using LLM as judge, BERTScore Average, and ROUGE average.}
    \label{tab:llm_judge_results}
\end{table*}

\subsection{Comparative Analysis and Clinical Implications}
MedGemma-27B with few-shot prompting achieved the highest deltaBLEU score (13.04), representing a 131\% improvement over the metadata ablation approach (5.70) as shown in Table 2. This superior performance can be attributed to the model's domain-specific medical training and optimal utilization of contextual examples. The consistency across human evaluation metrics (0.591-0.629) further validates this approach's clinical relevance.
However, when examining LLM-as-Judge evaluations (calculated as the average of DeepSeek-V3, GPT-4o, and Gemini scores) presented in Table 4, MedGemma-27B with metadata demonstrates superior performance (0.6342) compared to the few-shot approach without metadata (0.6185). This apparent contradiction with deltaBLEU scores suggests that while few-shot prompting excels in lexical similarity, the metadata-enhanced approach produces responses that are more clinically coherent and contextually appropriate according to expert-level language models. The metadata integration appears to provide structured clinical reasoning that resonates better with sophisticated evaluation frameworks, even though it may use different terminology than reference answers.
InternVL3-38B demonstrated intermediate deltaBLEU performance (9.92) despite utilizing a larger context window with 25 retrieved samples (Table \ref{tab:model_comparison_transposed}). Notably, the expanded few-shot context allows the model to access more familiar examples related to each query, resulting in improved performance across all evaluation dimensions including ROUGE (average: 0.566) and BERT score (0.645) compared to MedGemma-27B variants as detailed in Table \ref{tab:llm_judge_results}. The provision of 25 contextual examples enables better pattern recognition and clinical reasoning adaptation, though the general-purpose training limits its peak performance in specialized medical domains requiring precise wound-specific knowledge.
The metadata ablation study approach achieved the lowest deltaBLEU score (5.70) but maintains competitive ROUGE and BERT scores (Table \ref{tab:llm_judge_results}), suggesting that while lexical overlap may be reduced, semantic similarity and clinical relevance remain preserved. This indicates that the two-stage pipeline may suffer from error propagation during metadata prediction, and the confidence threshold mechanism (0.7) may have been overly conservative in integrating predicted clinical features, leading to more conservative but potentially more accurate clinical responses.
These findings highlight the trade-off between lexical similarity metrics and clinical appropriateness, emphasizing the importance of multi-faceted evaluation in medical AI systems where clinical accuracy often supersedes surface-level text matching.

\section*{Limitations}

The overall deltaBLEU scores across all approaches remain relatively modest, ranging from 5.70 to 13.04, which underscores the inherent complexity of medical visual question answering tasks, particularly in the specialized domain of wound care assessment. These moderate performance levels highlight fundamental challenges that must be addressed before such systems can provide meaningful clinical utility. Upon detailed examination of model outputs, we observed that while the systems demonstrate competency in identifying general wound characteristics and providing contextually appropriate clinical guidance, they frequently struggle with precise clinical terminology and specific wound classification. The models often generate responses that capture the general clinical context but may lack the precision required for definitive diagnostic support, similar to how they might correctly identify inflammatory characteristics while being less accurate in distinguishing between closely related wound types or infection stages that require different treatment protocols.
The variability in inter-annotator agreement scores reveals fundamental challenges inherent in the dataset itself, which directly impact model training and evaluation reliability. While features like wound type (IAA: 1.0) and tissue color (IAA: 0.97) show excellent agreement, the lower agreement for anatomical location (IAA: 0.81) and drainage amount (IAA: 0.86) suggests inherent subjectivity in clinical wound assessment that extends beyond simple annotation inconsistencies. This variability may reflect genuine clinical complexity, as wound characteristics often exist on continua rather than discrete categories, making it challenging for models to learn consistent decision boundaries. Furthermore, the dataset's scope may be limited in representing the full spectrum of wound presentations encountered in clinical practice, and the performance degradation observed when certain metadata categories are removed indicates potential dataset imbalances or insufficient representation of diverse wound presentations.
The methodological approaches employed in this study present several constraints that may have limited optimal performance. The two-stage pipeline approach in our metadata ablation study, while theoretically sound, appears to suffer from error propagation between metadata prediction and response generation phases, where inaccuracies in the initial metadata prediction cascade into the final response quality. The conservative confidence threshold (0.7) implemented may have been overly restrictive, limiting the integration of potentially valuable clinical insights and preventing the system from leveraging ambiguous but clinically relevant information. Additionally, the disparity in optimal context utilization across different models—requiring 5-shot prompting for MedGemma-27B versus 25-shot prompting for InternVL3-38B—suggests that current few-shot learning strategies are highly model-dependent and may require more systematic optimization approaches tailored to specific architectural characteristics.
The gap between semantic similarity metrics and clinical accuracy presents a significant concern for practical deployment. While BERTScore consistency indicates that models maintain coherent medical discourse, the modest deltaBLEU scores suggest they may not achieve the diagnostic precision necessary for clinical decision support. This discrepancy is particularly problematic in wound care, where treatment decisions often hinge on subtle clinical distinctions that our current approaches may not adequately capture. The models' tendency to provide generally appropriate clinical context while missing specific diagnostic details could potentially lead to suboptimal treatment recommendations or delayed appropriate interventions in real clinical settings.
Current evaluation frameworks may not fully capture the complexities of clinical utility and decision-making processes. The reliance on text-based similarity metrics, while providing standardized comparison methods, may not adequately reflect the nuanced clinical reasoning required for effective wound care assessment. The evaluation approach does not account for the hierarchical importance of different types of clinical information—where certain diagnostic errors may have more severe clinical consequences than others—nor does it assess the models' ability to appropriately express uncertainty when faced with ambiguous presentations. Additionally, the absence of longitudinal assessment data limits our understanding of how these systems might perform in tracking wound healing progression or adapting recommendations based on treatment responses, which are critical components of comprehensive wound care management in clinical practice.

\section*{Conclusion}
This study evaluated three distinct approaches for wound care visual question answering, revealing significant challenges and opportunities in medical multimodal AI systems. MedGemma-27B with few-shot prompting achieved the highest performance (deltaBLEU: 13.04), demonstrating the value of domain-specific medical training over general-purpose multimodal architectures. The metadata ablation study identified anatomical location, wound type, and tissue color as critical features for wound assessment, with their removal causing substantial performance degradation. However, the overall modest deltaBLEU scores (5.70-13.04) underscore the inherent complexity of medical visual question answering tasks and highlight the substantial improvements required before clinical deployment.
The variability in inter-annotator agreement scores across wound attributes reflects genuine clinical complexity rather than simple annotation inconsistencies, emphasizing the subjective nature of certain wound characteristics. While models demonstrated competency in generating contextually appropriate clinical guidance, they frequently struggled with precise diagnostic terminology and specific wound classification—critical requirements for effective clinical decision support.
Future research should prioritize hybrid architectures that combine multimodal reasoning capabilities with specialized medical knowledge, develop more sophisticated uncertainty quantification methods, and establish evaluation frameworks that better align with clinical decision-making processes. Enhanced datasets incorporating diverse wound presentations and longitudinal treatment data, coupled with comprehensive clinical validation studies, are essential steps toward developing systems that can meaningfully contribute to healthcare practice. The gap between current performance and clinical requirements necessitates continued interdisciplinary collaboration between AI researchers and healthcare professionals to address these fundamental challenges.

 \section*{Future work}
Current deltaBLEU scores (5.70-13.04) suggest considerable enhancement potential compared to theoretical maximums. Upcoming research directions include optimizing few-shot sample selection beyond semantic similarity through diversity-based approaches and adaptive context sizing. Implementing ensemble metadata classifiers for specific clinical features could minimize error propagation while improving integration confidence thresholds. Developing fine-tuned specialized medical embeddings through augmented wound imagery from training datasets would enhance both metadata prediction and similarity matching. Additionally, creating generalized Visual Question Answering frameworks for comprehensive wound care, rather than topic-specific models, would improve scalability across diverse clinical scenarios. These advancements collectively address performance limitations while establishing foundations for robust healthcare applications.
\section*{Acknowledgments}
We extend our sincere thanks to EXL Health AI Lab for their support and computing resources. We also appreciate the efforts of our colleagues who contributed to discussions and provided valuable assistance during the course of this work. Finally, we acknowledge the organizers for their efforts in hosting this interesting and challenging competition.


\bibliography{custom}

\appendix
\section{Prompts used in Approach}

\subsection{LLM as Judge Prompt}

\textbf{SYSTEM:} You are a helpful medical assistant.

\textbf{USER:} Given a patient QUERY, and a list of REFERENCE RESPONSES, please evaluate a CANDIDATE RESPONSE using a three-step rating described below.

Rating: 0 - CANDIDATE RESPONSE is incomplete and may contain medically incorrect advice.

Rating: 0.5 - CANDIDATE RESPONSE is incomplete but has partially correct medical advice.

Rating: 1.0 - CANDIDATE RESPONSE is complete and has medically correct advice.

The REFERENCE RESPONSES represent answers given by domain experts and can be used as references for evaluation.

QUERY:

REFERENCE RESPONSES:

CANDIDATE RESPONSE:

RATING:

\subsection{Mined few shot approach prompt}

\textbf{System prompt:} You are a clinical AI assistant with expertise in wound care and infection prevention, responsible for answering patient queries.

\textbf{Prompt template:}

Given the patient's query and associated wound images, your task is to:
\begin{itemize}
    \item Analyze the query and images together
    \item Identify likely wound condition or stage
    \item Suggest appropriate wound care steps (cleaning, dressing, follow-up)
    \item Warn if urgent medical attention is required
    \item Keep the tone concise, clinical, and avoid unnecessary details
\end{itemize}

EXAMPLES:

[few shot examples]

Now, based on the format of the above examples, generate a response for the following query. Strictly follow the example style and do not include any headings in your response.

Patient Query:

Query Title: [query title]

Query Content: [query content]

\subsection{Metadata study approach prompt}

\textbf{System prompt:} You are an expert wound care assistant specializing in interpreting wound images and providing concise, medically sound advice. Given a clinical query and one or more wound images, your job is to deliver short, accurate answers based on visible findings and basic wound care principles. Use clinical reasoning to interpret visual cues (e.g., redness, scabbing, swelling, sutures, necrosis, granulation tissue). You are a medical wound-care assistant. Provide clinically accurate and safe guidance based on the query, wound images, and metadata. Your responses should be medically helpful, crisp, and no longer than 2-3 sentences. Avoid lengthy explanations or disclaimers. If urgent care is required, clearly recommend it. Otherwise, suggest simple, evidence-based wound care actions.

\textbf{Classification system prompt:} You are a wound-care classification assistant. Return wound metadata with calibrated confidence scores. For each field: Choose ONLY from allowed values. Provide a numeric confidence score in [0,1] (0=very unsure, 1=highly certain). For anatomic locations (multi-label), include each predicted body location as an object with label + score. Only include locations you believe are present (score greater than 0.25). Sort them by descending score. If you are unsure for a field, output a best guess with a low score; do NOT invent values outside the allowed lists. Compute an overall certainty = average of all individual field confidences (use mean of chosen location scores for anatomic locations). Output STRICT valid JSON only, matching this schema: predictions with anatomic locations, wound type, tissue color, drainage type; scores with anatomic locations confidence, wound type confidence, tissue color confidence, drainage type confidence, overall certainty. No extra text.

\textbf{User prompt classification:} CLASSIFY THE FOLLOWING IMAGES AND QUERY. Allowed values: [class values]. FEW SHOT EXAMPLES: [few shot examples]. Patient Query: Title: [query title]. Content: [query content].

\textbf{User prompt response:} You are a highly skilled clinical wound-care assistant trained to provide safe, concise, and medically sound advice. You will receive: The patient's wound query (title and content). Predicted wound metadata across 4 key wound-related categories. Confidence scores (0 to 1) for each metadata field. Your task is to: (1) Generate an initial clinical response based solely on the patient's query (title + content), without referring to the metadata. (2) Reflect on the predicted metadata and its confidence scores. Then, evaluate whether the initial response can be improved using this structured information. (3) If the metadata confidence is high (greater than 0.7), refine your response using these metadata details to make it more targeted and informative. (4) If confidence is low (less than 0.7), do not make firm assumptions based on those fields. Instead, express clinical caution or recommend seeking professional guidance. (5) Pay particular attention to the wound type and wound thickness fields. Avoid overconfident guidance when these have low confidence. (6) Ensure that metadata like anatomic locations and tissue color (when reliable) inform and personalize your response. (7) Keep the final response clinically sound, concise (50 tokens or fewer), and empathetic. Patient Query: Title: [query title]. Content: [query content]. Predicted Metadata: [metadata]. Only return the final response without any additional text and within 50 tokens.

\end{document}